%% file: main.tex
\pdfoutput=1
\documentclass[11pt]{article}

\usepackage{emnlp2021}

\usepackage{times}
\usepackage{latexsym}

\usepackage[T1]{fontenc}
\usepackage[utf8]{inputenc}

\usepackage{microtype}

\usepackage{graphicx}
\usepackage{subfig} 

\input{settings.tex}

\input{math_commands.tex}

\algdef{SE}[DOWHILE]{Do}{doWhile}{\algorithmicdo}[1]{\algorithmicwhile\ #1}%

    \makeatletter
\def\@fnsymbol#1{\ensuremath{\ifcase#1\or *\or \diamond\or
   \mathsection\or \mathparagraph\or \|\or **\or \dagger\dagger
   \or \ddagger\ddagger \else\@ctrerr\fi}}
    \makeatother

\newcommand\ours{\textsc{VoCap}}
\newcommand\joint{\textsc{Joint}}

\title{Allocating Large Vocabulary Capacity for \\ Cross-lingual Language Model Pre-training}

\author{Bo Zheng$^\dag$\thanks{\ \  Contribution during internship at Microsoft Research.},~~Li Dong$^\ddag$,~~Shaohan Huang$^\ddag$,\\
\textbf{Saksham Singhal}$^{\ddag}$\textbf{,}~~\textbf{Wanxiang Che}$^{\dag}  $\textbf{,}~~\textbf{Ting Liu}$^\dag$\textbf{,}~~\textbf{Xia Song}$^\ddag$\textbf{,}~~\textbf{Furu Wei}$^\ddag$\\
$^\dag$Harbin Institute of Technology \\
$^\ddag$Microsoft Corporation \\
\texttt{\{bzheng,car,tliu\}@ir.hit.edu.cn} \\
\texttt{\{lidong1,shaohanh,saksingh,xiaso,fuwei\}@microsoft.com} \\}

\begin{document}
\maketitle
\begin{abstract}
Compared to monolingual models, cross-lingual models usually require a more expressive vocabulary to represent all languages adequately. We find that many languages are under-represented in recent cross-lingual language models due to the limited vocabulary capacity. To this end, we propose an algorithm \ours{} to determine the desired vocabulary capacity of each language. However, increasing the vocabulary size significantly slows down the pre-training speed. In order to address the issues, we propose $k$-NN-based target sampling to accelerate the expensive softmax. Our experiments show that the multilingual vocabulary learned with \ours{} benefits cross-lingual language model pre-training. Moreover, $k$-NN-based target sampling mitigates the side-effects of increasing the vocabulary size while achieving comparable performance and faster pre-training speed. The code and the pretrained multilingual vocabularies are available at \url{https://github.com/bozheng-hit/VoCapXLM}.
\end{abstract}

\section{Introduction}
\label{sec:intro}

Pretrained cross-lingual language models~\citep{xlm, xlmr, infoxlm, mt5} have recently shown great success in improving cross-lingual transferability.
These models encode texts from different languages into universal representations with a shared multilingual vocabulary and a shared Transformer encoder~\citep{transformer}. By pre-training cross-lingual language models on the large-scale multilingual corpus, the models achieve state-of-the-art performance on various downstream tasks, e.g., cross-lingual question answering and cross-lingual sentence classification.

Although the Transformer architecture used in most pretrained monolingual and cross-lingual language models are almost identical, the vocabularies are quite different.
The vocabulary sizes in existing pretrained monolingual language models typically range from 30K to 60K subword units~\citep{bert, roberta,unilm,unilmv2}. 
Meanwhile, state-of-the-art pretrained cross-lingual language models use the shared multilingual vocabulary of 250K subword units to represent more than 100 languages~\citep{xlmr, infoxlm, mt5}.
Although some subword units are shared across languages, no more than 2.5K language-specific subword units on average are allocated for each language, which is still relatively small.
Besides, the multilingual vocabulary is trained on the combined multilingual corpus with subword segmentation algorithms like BPE~\cite{bpe} and unigram language model~\cite{DBLP:conf/acl/Kudo18}.
During vocabulary construction, these algorithms tend to select more subword units shared across languages with common scripts like Latin and Cyrillic~\cite{DBLP:conf/emnlp/ChungGTR20}, but have a lower chance to select language-specific subword units. 
It is hard to determine how much vocabulary capacity a particular language requires and whether the shared multilingual vocabulary has allocated enough vocabulary capacity to represent the language.

In this paper, we propose \ours{}, an algorithm to allocate large vocabulary for cross-lingual language model by separately evaluating the required vocabulary capacity of each language.
First, we use the \textit{average log probability} (ALP) to evaluate the ability of a vocabulary to represent a particular language. 
We find that ALP is highly correlated to the downstream task performance, and we use it as an indicator to allocate language-specific vocabulary capacity.
In addition, the language-specific pre-training corpus size should also be considered since the pretrained model can only learn limited knowledge from low-resource languages where the pre-training data is scarce. Therefore, allocating too much vocabulary capacity for low-resource languages is inefficient. 
\ours{} leverages both ALP and pre-training corpus size to evaluate the required vocabulary capacity of each language.
We finally allocate a multilingual vocabulary with 500K subword units with \ours{} and show it can significantly improve the model performance.

However, increasing the vocabulary size has two practical drawbacks: slow pre-training speed and heavy model size. 
To address the pre-training speed issue, we propose $k$-NN-based target sampling, an approximate algorithm to improve the computing efficiency in the expensive softmax caused by the large vocabulary.
We pre-train the model with a small subset of the entire vocabulary constructed with $k$ nearest neighbors of the target words in current mini-batch data, evaluated with the inner product of subword embeddings.
As for the model size, we halve the embedding dimension and draw a different conclusion from~\citet{xlmr} that increasing vocabulary from 250K to 500K with a fixed capacity model can also improve the performance.

Our contributions are summarized as follows: 
\begin{itemize}
\item We propose \ours{}, an algorithm to allocate appropriate vocabulary capacity for each language in the shared multilingual vocabulary of cross-lingual language models.
\item We propose $k$-NN-based target sampling, a softmax approximation algorithm to improve the computing efficiency during cross-lingual language model pre-training. 
\item We evaluate our methods on the XTREME benchmark~\citep{xtreme}, including three different tasks on seven datasets. 
Experiments show that \ours{} consistently outperforms previous vocabulary construction methods. 
Meanwhile, our $k$-NN-based target sampling enables effective acceleration while achieving comparable performance.
\end{itemize}

\begin{figure*}[t]
\centering
\begin{minipage}[t]{0.48\textwidth}
\centering
\includegraphics[width=8cm,trim={0.5cm 0cm 1cm 1cm}]{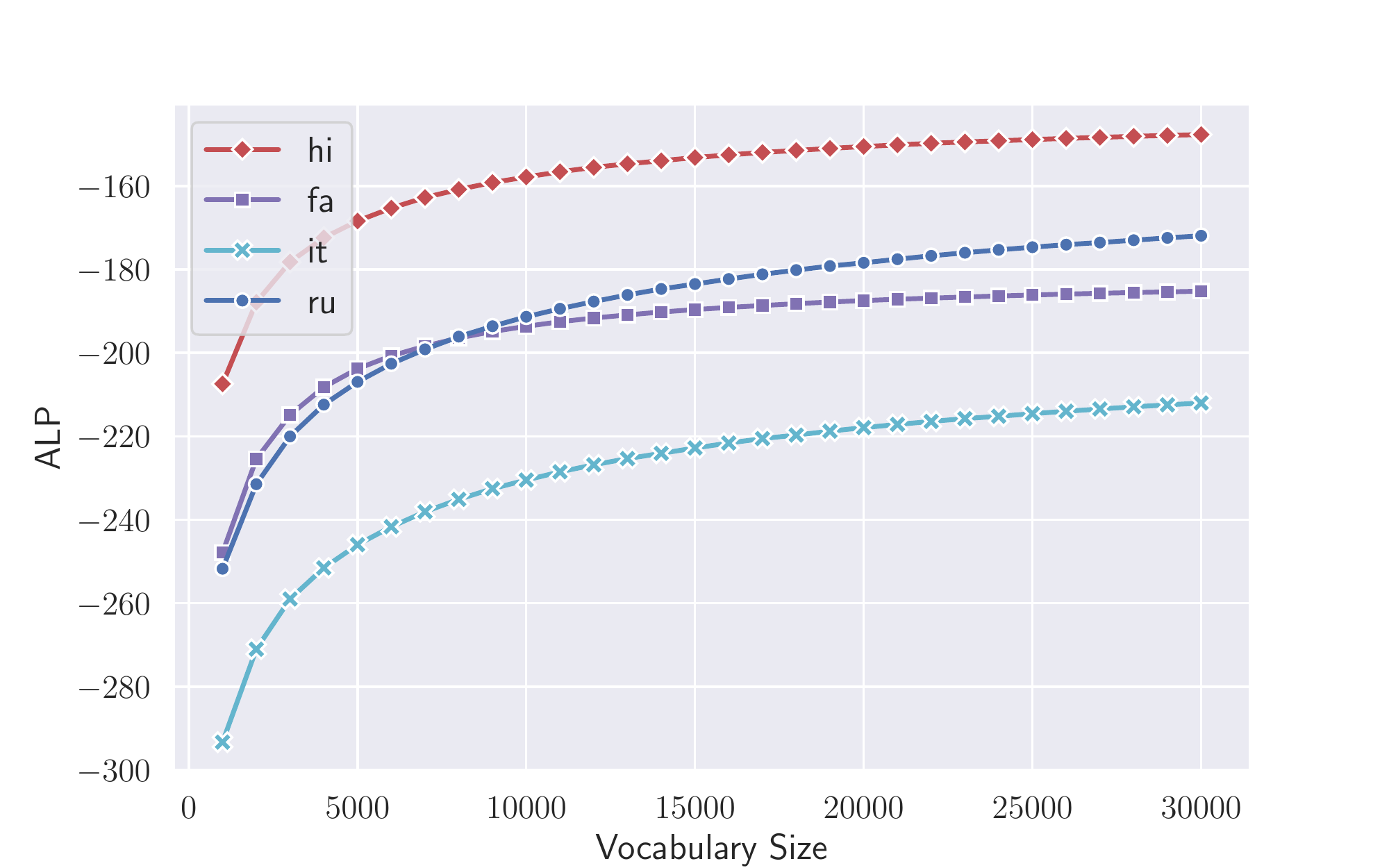}
\caption{ALP of different monolingual vocabularies with different vocabulary sizes. }
\label{fig:alp}
\end{minipage}
\quad
\begin{minipage}[t]{0.48\textwidth}
\centering
\includegraphics[width=8cm,trim={0.5cm 0cm 1cm 1cm}]{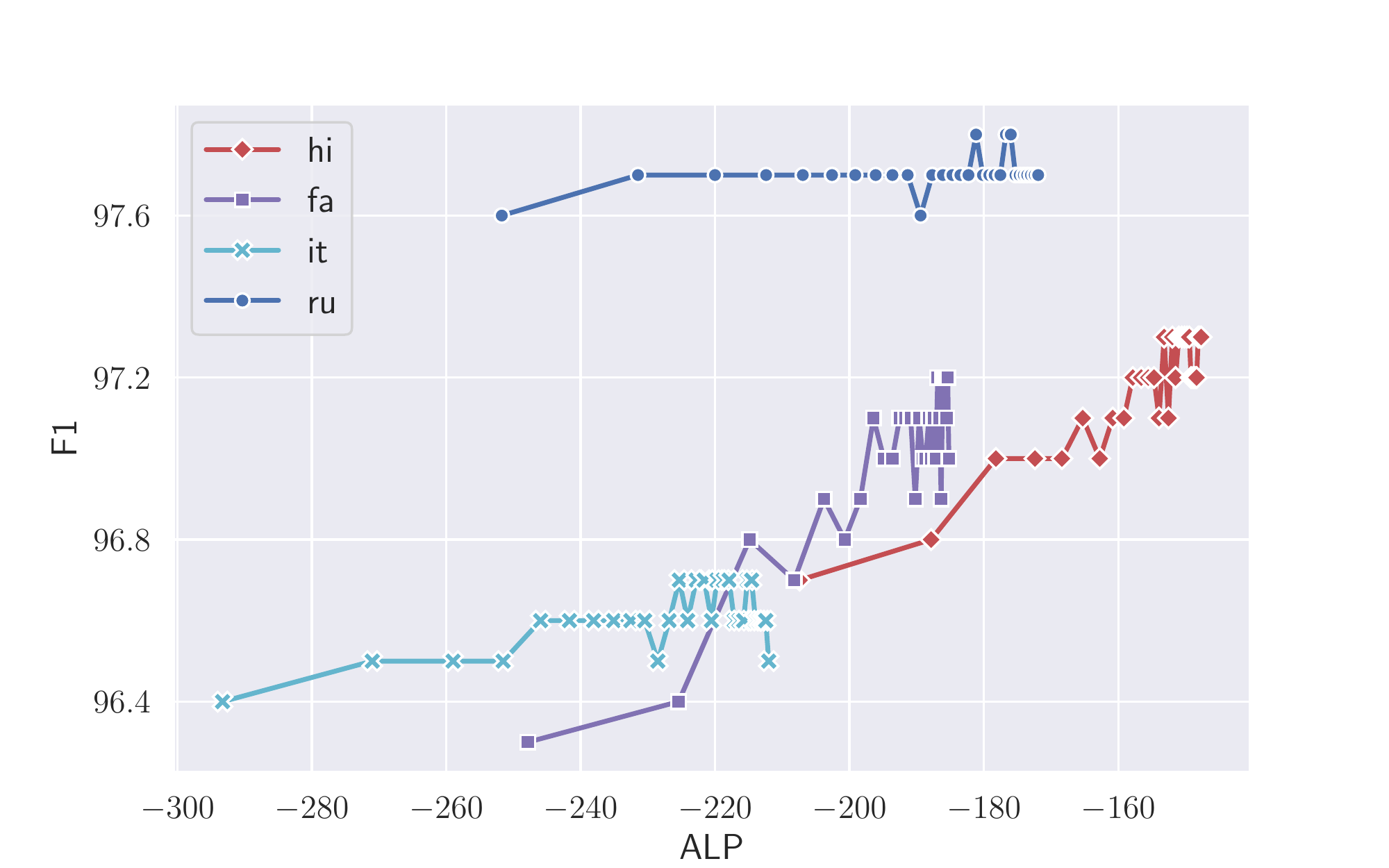}
\caption{F1 score on POS task with different vocabularies versus their ALP on the monolingual corpus.}
\label{fig:udpos}
\end{minipage}
\begin{minipage}[t]{0.48\textwidth}
\centering
\includegraphics[width=8cm,trim={0.5cm 0cm 1cm 0cm}]{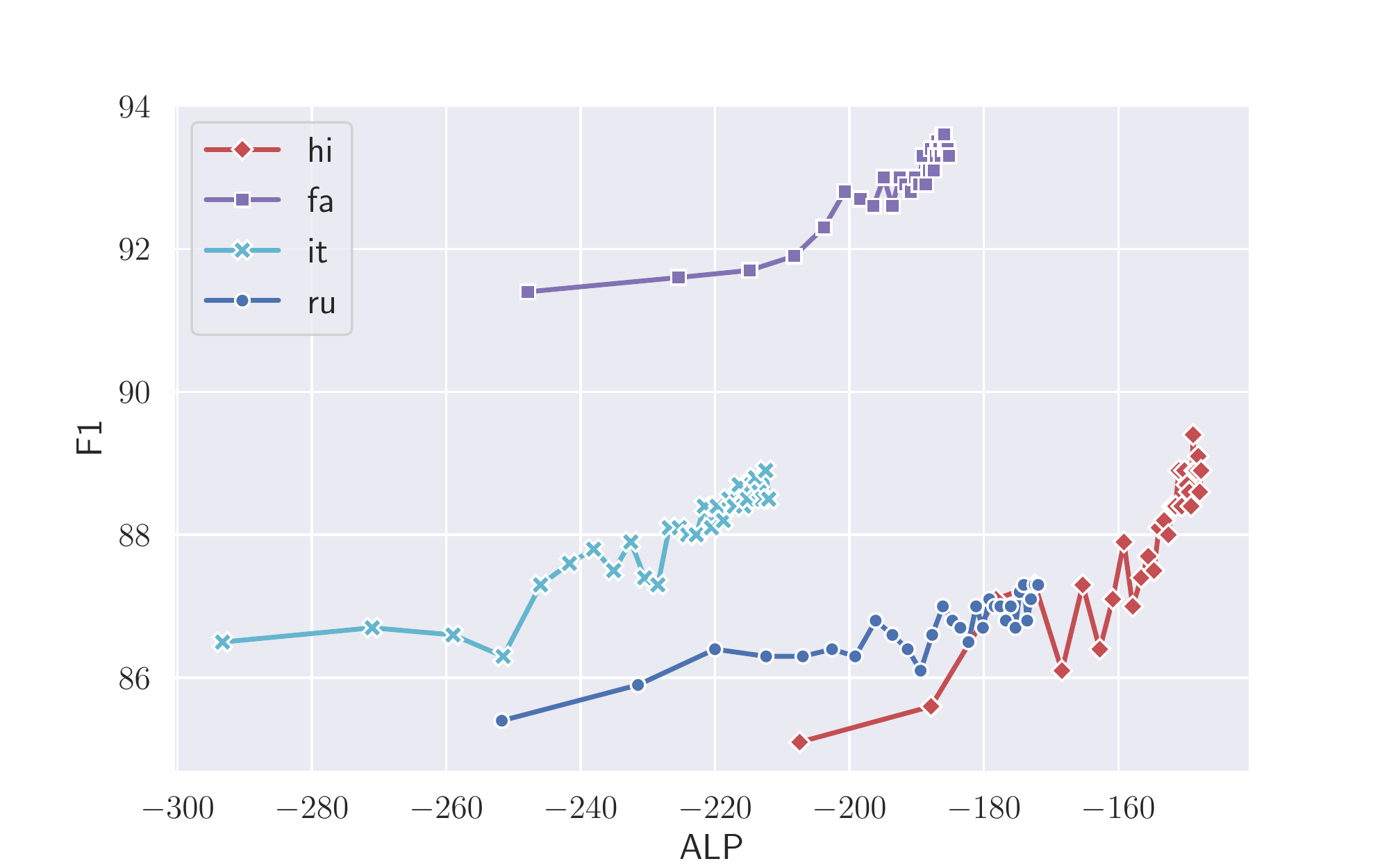}
\caption{F1 score on NER task with different vocabularies versus their ALP on the monolingual corpus. }
\label{fig:panx}
\end{minipage}
\quad
\begin{minipage}[t]{0.48\textwidth}
\centering
\includegraphics[width=8cm,trim={0.5cm 0cm 1cm 0cm}]{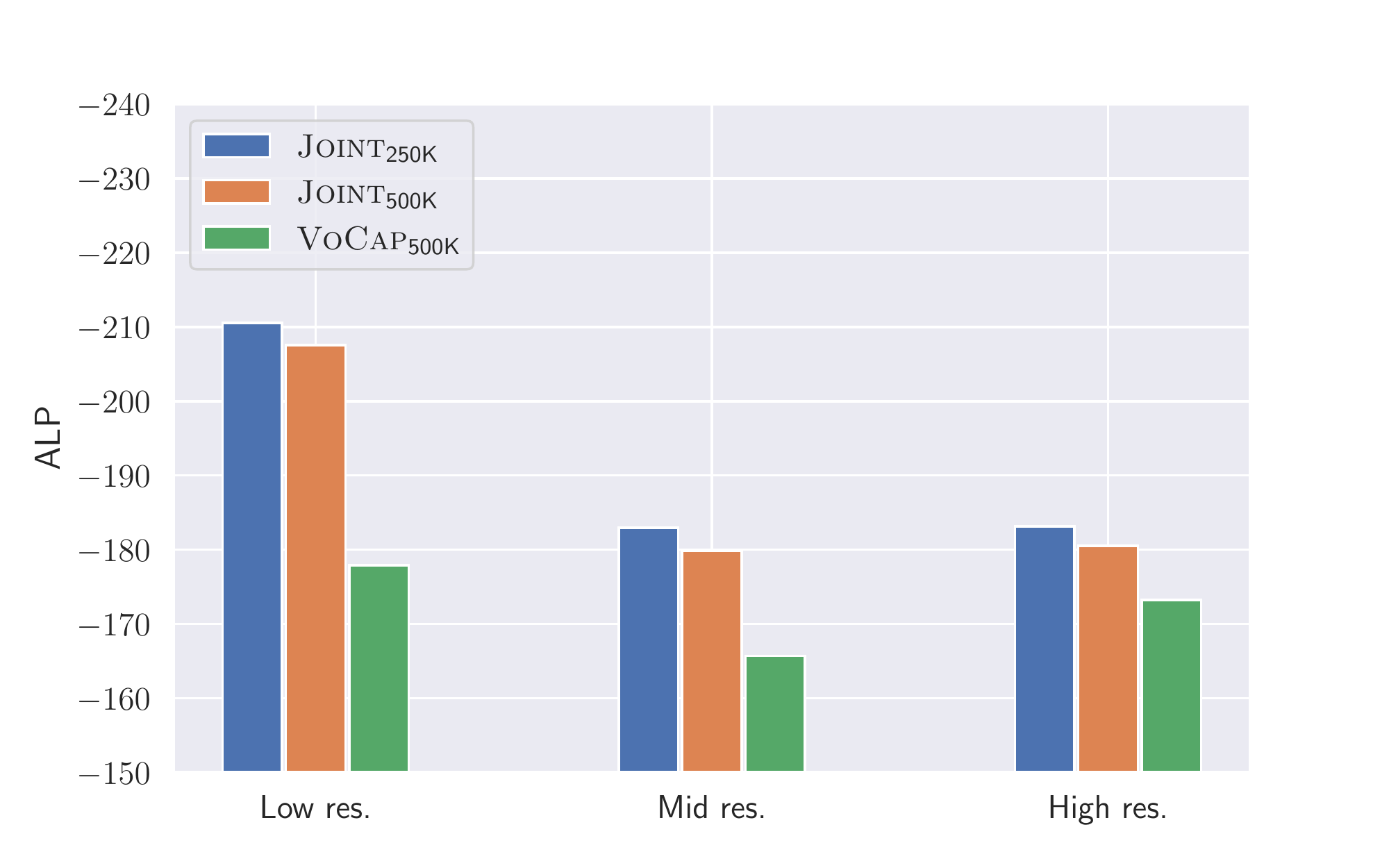}
\caption{
Comparison of vocabulary capacity of different-resourced languages. Shorter bars indicate larger vocabulary capacity.
}
\label{fig:res-alp}
\end{minipage}
\end{figure*}

\section{\ours{}: Language-Specific Vocabulary Capacity Allocation}
\label{sec:vocap}
We attribute the main factors that affect the performance of a particular language in a cross-lingual language model to language-specific pre-training corpus size and vocabulary capacity.
While previous work adjusts pre-training corpus size with an exponentially smoothed sampling distribution~\citep{xlm,xlmr}, few existing works have explored the effect of the language-specific vocabulary capacity in pretrained cross-lingual language models. 

In this section, we first investigate the correlation between the language-specific vocabulary capacity and downstream task performance through experiments.
Then we introduce our proposed multilingual vocabulary allocation algorithm \ours{}.

\subsection{Investigating Language-Specific Vocabulary Capacity}
We start by introducing \textit{average log probability} (ALP) to quantify the language-specific vocabulary capacity in the shared multilingual vocabulary for a specific language.\footnote{For brevity and consistency, we refer to the parameterized tokenizer also as vocabulary, e.g., SentencePiece model.} Given a monolingual corpus composed of sentences $\mathcal{D}_{i}=\{s_1,...,s_{|\mathcal{D}_{i}|} \}$ from the $i$-th language and tokenized with vocabulary $V$, the average log probability is defined as follows:
\begin{align}
\textrm{ALP}(\mathcal{D}_{i}, V) = \frac{1}{|\mathcal{D}_{i}|}\sum_{j=1}^{|\mathcal{D}_{i}|}\sum_{k=1}^{|s_j|} \log p_{uni}(s_j^k)
\label{eq:alp}
\end{align}
where $s_j^k$ is the $k$-th subword of the sentence $s_j$, and $p_{uni}(\cdot)$ is the unigram distribution counted on the monolingual corpus $\mathcal{D}_{i}$. 
It is difficult to count the language-specific subword units in multilingual vocabularies since the raw text contains a lot of code-switched data. 
By contrast, ALP is a more convenient indicator of language-specific vocabulary capacity and it is penalized by the subword units with low-frequency.

To investigate the impact of language-specific vocabulary capacity, we first learn monolingual vocabularies in different sizes to obtain vocabularies with different ALP, i.e., language-specific vocabulary capacity. 
Then we conduct pre-training with these monolingual vocabularies on their corresponding monolingual corpora. 
Finally, we evaluate these monolingual models on downstream tasks and study the correlation between language-specific vocabulary capacity and downstream task performance.

\subsubsection{Setup}
To alleviate the bias from the languages' characteristics, we first select four languages with different pre-training corpus sizes from different language families, which are Hindi (hi), Persian (fa), Italian (it), Russian (ru). 
We first learn thirty monolingual vocabularies for each language on the corresponding monolingual corpus, with vocabulary size ranging from 1K to 30K.
Then we pretrain monolingual language models with the corresponding monolingual vocabularies.
We evaluate these pretrained models on two downstream tasks:  NER~\cite{panx} and POS~\cite{udpos} from the XTREME benchmark since there is annotated task data for a large number of languages. 
The vocabularies are learned on the reconstructed CommonCrawl corpus~\citep{infoxlm, xlmr} using SentencePiece~\citep{sentencepiece} with the unigram language model~\citep{DBLP:conf/acl/Kudo18}. The unigram distributions are also counted on the CommonCrawl corpus.
The Wikipedia corpus is used for all pre-training experiments in this paper since it is easier to run experiments due to its smaller size. 
More details about the pre-training data can be found in the appendix.

\subsubsection{Observations}
\label{sec:obs}

\paragraph{Increasing vocabulary size affects ALP of different languages in varying degrees.} 
In Figure~\ref{fig:alp}, we show the correlation between vocabulary size and ALP of four different languages. 
We observe the ALP varies across different languages, mainly because ALP correlates with the lexicon granularity of the language, i.e., the average number of tokens per sentence.
Besides, when the vocabulary size is larger than 10,000, the gains of increasing monolingual vocabulary size in hi and fa are less than it and ru. We attribute it to that hi and fa does not have extensive compoundings. 
Another observation is that for each language, every time we increase the vocabulary size by 1K, the increment in ALP is monotonically decreasing. 

\paragraph{ALP correlates positively with downstream task performance.} 
In Figure~\ref{fig:udpos} and Figure~\ref{fig:panx}, 
we illustrate downstream task performance of models pretrained with monolingual vocabularies on corresponding monolingual corpora. 
We observe that ALP correlates positively with downstream task performance, making language-specific ALP a valid indicator to allocate multilingual vocabulary.
Another natural option to allocate multilingual vocabulary is directly using monolingual vocabulary size to indicate language-specific vocabulary capacity.
We compare ALP against vocabulary size and observe that ALP correlates better than vocabulary size with the downstream task performance. Besides, ALP reflects the language-specific characteristics, while vocabulary size does not. The detailed comparison is shown in the appendix.

\subsection{Allocating Multilingual Vocabulary with \ours{}
}

Based on the observations in Section~\ref{sec:obs}, we first give the implementation of our proposed vocabulary allocation algorithm \ours{}. Then we compare the multilingual vocabulary learned with \ours{} and directly learned with SentencePiece on the multilingual corpus.

\subsubsection{\ours{} Implementation}
\label{sec:vocap-imp}

\begin{algorithm}[t]
\small
\caption{\small Allocating Multilingual Vocabulary with \ours{}}
\label{alg:vocap}
\begin{algorithmic}[1]
\Require size of target multilingual vocabulary $T$; monolingual vocabularies of $N$ languages $\{V^{i}_{t_{i}}\}_{i=1}^{N}$; monolingual corpus of $N$ languages $\{D_i\}_{i=1}^{N}$
\Ensure multilingual vocabulary $V$
\For{$i \gets 1$ \textbf{to} $N$}
    \For{$j \gets 1$ \textbf{to} $50$}
        \State $a_{i,j \times 1000} \gets \textrm{ALP}(D_i, V^i_{j \times 1000})$
    \EndFor
    \State $t_{i} \gets 0$
    \State $a_{i,0} \gets -\infty$
\EndFor

\Do
    \State $j \gets 0$
    \State $\delta \gets 0$
    \For{$i \gets 1$ \textbf{to} $N$} 
        \If{$\delta < a_{i, t_i+1000} - a_{i, t_i}$}
            \State $\delta \gets a_{i, t_i+1000} - a_{i, t_i}$
            \State $j \gets i$
        \EndIf
    \EndFor
    \State $t_j \gets t_j + 1000$
    \State $V \gets |\bigcup_{i=1}^{N}V^{i}_{t_i}|$
\doWhile{$|V| < T$}
\If{$|V| > T$}
\State Clip the size of $V$ to $T$
\EndIf
\end{algorithmic}
\end{algorithm}

We formulate the vocabulary construction of \ours{} as the problem of finding the optimal way to allocate language-specific vocabulary size to each language, such that the overall ALP of all languages is maximized.
In addition to language-specific vocabulary capacity measured with ALP from \eqform{eq:alp}, the language-specific pre-training corpus size also affects the downstream task performance.
Considering the two factors, the procedure of \ours{} can be formulated as follows:
\begin{align}
\underset{t_{1},...,t_{N}}{\mathrm{argmax}} \sum_{i=1}^{N}q_i^{\beta}\mathrm{ALP}(D_i, V^{i}_{t_i})
~~s.t.~~|\bigcup_{i=1}^{N}V^{i}_{t_i}| = T
\label{eq:vocap}
\end{align}
where $t_{i} \in \{x \times 1000 \mid x \leq 50, x \in N^+\}$ is the number of subword units allocated to the $i$-th language,\footnote{Since the cost of learning monolingual vocabularies with arbitrary sizes and getting the corresponding ALP is unaffordable, we learn monolingual vocabularies with vocabulary size range from 1K to 50K at intervals of 1K. } $\beta$ is a rescaling factor, $V^{i}_{t_{i}}$ is the vocabulary of the $i$-th language with $t_{i}$ subword units, $T$ is the size of the target multilingual vocabulary, and $q_{i}$ is the probability of sampling training instances from $i$-th language during pre-training~\citep{xlm, xlmr}:
\begin{align}
q_i = \frac{f_{i}^{\alpha}}{\sum_{j=1}^{N}f_{j}^{\alpha}}~~\text{with}~~f_i = \frac{n_i}{\sum_{k=1}^{N}n_k}
\label{eq:sampling}
\end{align}
where $n_i$ is the number of instances in the $i$-th language,
$\alpha$ is a rescaling factor used to alleviate the bias towards high-resource languages. 
Since the increment in ALP when increasing the vocabulary size by a certain number is monotonically decreasing, \eqform{eq:vocap} can be solved with the greedy algorithm in Algorithm~\ref{alg:vocap}.

\subsubsection{Intrinsic Analysis}
We compare the multilingual vocabulary learned with \ours{} and directly learned with SentencePiece on the multilingual corpus.
The multilingual corpus to learn vocabularies in this paper is the concatenation of sentences sampled randomly from the monolingual corpora. Sentences from the $i$-th language is sampled with probability $q_i$ from \eqform{eq:sampling} and use $\alpha=0.7$. We filter languages with corpus size larger than 0.1 GB, resulting in 86 languages.

We evaluate the multilingual vocabularies with their ALP on each language's monolingual corpus, and show results of different-resourced languages in Figure~\ref{fig:res-alp}. We refer to languages with less than 1GB and more than 10GB pre-training corpus in the reconstructed CommonCrawl as low-resource and high-resource languages, respectively, otherwise mid-resource languages. 
When directly learning vocabulary on the multilingual corpus using SentencePiece, the vocabulary with 500K subword units ($\joint{}_{\text{500K}}$) only has a negligible improvement compared to the vocabulary with 250K subword units ($\joint{}_{\text{250K}}$).
Meanwhile, our method ($\ours{}_{\text{500K}}$) consistently outperforms $\joint{}_{\text{500K}}$ in different-resourced languages, especially in mid and low-resource languages. The statistics of the allocated vocabulary size for each language in $\ours{}_{\text{500K}}$ are shown in the appendix.

\section{Accelerate Large-Vocabulary Language Model Pre-Training}

Although extending the multilingual vocabulary benefits cross-lingual language models, pre-training with such large vocabularies brings two practical issues: slow pre-training speed and heavy model size. 
To tackle the issues, we first introduce our $k$-NN-based target sampling in Section~\ref{sec:kts}, which is a softmax approximation algorithm to improve computing efficiency. 
Then we describe how we reallocate the model parameters to keep the model size fixed in Section~\ref{sec:mb}.

\subsection{$k$-NN-Based Target Sampling}
\label{sec:kts}
To reduce the expensive computation cost of the softmax function, we propose $k$-NN-based target sampling to approximate the expensive softmax. 
The original masked language modeling objective minimizes the cross-entropy loss for every masked subword $w_i$ on the extensive multilingual vocabulary $V$. 
The proposed $k$-NN-based target sampling instead uses a smaller vocabulary subset $V'$. The approximation of the masked language modeling loss for the masked subword $w_i$ is defined as follows:
\begin{align}
\mathcal{L}(w_i) =
-\mathrm{log}\frac{\exp(h^{\mathrm{T}}v_{w_i} + b_{w_{i}})}{\sum_{w_{j} \in V'}\exp(h^{\mathrm{T}}v_{w_j} + b_{w_j})}
\label{eq:loss}
\end{align}
where $h$ is the corresponding output vector of the penultimate network layer, i.e., the output vector of the Transformer encoder, $v_{w_{i}}$ is the embedding of the subword unit $w_{i}$, and $b_{w_{i}}$ is a bias term.
We formulate the construction of the vocabulary subset $V'$ as follows:
\begin{align}
V' &= \bigcup_{w_i \in \mathcal{W}} \mathcal{I}_{k}(w_i) \\
\mathcal{I}_{k}(w_i) &= \mathrm{top\text{-}k}(\{ v_{w_i}^{\mathrm{T}}v_{w_j} \mid w_j \in V \})
\label{eq:subvocab}
\end{align}
where $\mathcal{W}$ denotes the set of target masked subword units in the current mini-batch, and $\mathcal{I}_{k}(w_i)$ denotes the $k$ most similar subwords measured with the inner product of the subword embedding $v_{w_i}$ and $v_{w_j}$. 

However, retrieving $\mathcal{I}_k(w_i)$ at every training step for every subword unit $w_i \in \mathcal{W}$ requires as much computation cost as softmax, which is unaffordable.
As an alternative, we compute $\mathcal{I}_k(w_i)$ for every subword $w_i \in V$ according to the current subword embeddings every $n$ training steps and replace the previous version of $\mathcal{I}_k(w_i)$ with the new one. We determine the value of $n$ such that $|V| \ll n \times |\mathcal{W}|$.
We illustrate the pre-training procedure with $k$-NN-based target sampling in Algorithm~\ref{alg:kts}.

\begin{algorithm}[t]
\small
\caption{\small Pre-training with $k$-NN-based target sampling}
\label{alg:kts}
\begin{algorithmic}[1]
\Require multilingual corpus $\train_\text{m}$; size $k$ of $k$-NN-based target sampling; multilingual vocabulary $V$; learning rate $\tau$
\Ensure model parameters $\vtheta$

\While{not converged}
\State Sample $n$ mini-batches $\{\mathcal{X}^{(t)},  \mathcal{W}^{(t)}\}_{t=1}^{n}$ $\sim$ $\train_\text{m}$ \Comment{\textit{\color{gray} $\mathcal{X}^{(t)}$ is a mini-batch of monolingual text, and $\mathcal{W}^{(t)}$ is the set of masked subwords.}}
\State Update $\mathcal{I}_{k}(w_i)$ for every $w_i \in V$
\For{$t \gets 1$ \textbf{to} $n$} \Comment{\textit{\color{gray} Train the model for $n$ steps.}}
\State $V' \gets \bigcup_{w_i \in \mathcal{W}^{(t)}} \mathcal{I}_{k}(w_i)$ 
\State $\vg \gets \sum_{w_i \in \mathcal{W}^{(t)}}\nabla_\vtheta \Ls(w_i)$
\State $\vtheta \gets \vtheta - \tau \vg$
\EndFor
\EndWhile
\end{algorithmic}
\end{algorithm}

From a practical point of view under the cross-lingual setting, the previous sampling-based softmax approximation methods either sample subwords from recent mini-batches or samples subwords from unigram distribution, 
the task becomes simpler since a considerable part of the subword samples is from different languages. 
Meanwhile, our $k$-NN-based target sampling uses subwords with similar representations like synonyms, which enforces the model focus on discriminating the ground-truth subword from a set of noise samples that are not easy to distinguish.
When using an approximate algorithm, the key point is to remain the difficult part of the original masked language modeling objective as much as possible.

\begin{table*}[t]
\centering
\small
\setlength{\tabcolsep}{1.0mm}
\begin{tabular}{lccccccccccc}
\toprule
\multirow{2}{*}{\bf Model} & \multirow{2}{*}{\bf \# Params} & \multirow{2}{*}{\bf Speed} & \multicolumn{2}{c}{\bf Pair Sentence} & \multicolumn{2}{c}{\bf Structure Prediction} & \multicolumn{3}{c}{\bf Question Answering} &  \\
& & & XNLI & PAWS-X & POS & NER & XQuAD & MLQA & TyDiQA \\ \midrule
 & & & Acc. & Acc. & F1 & F1 & F1/EM & F1/EM & F1/EM & Avg. \\ \midrule
$\text{XLM-R}_{\text{250K}}$ & 265M & 1.00x & 68.7 & 82.6 & 72.1 & 60.6 & 63.4/47.4 & 57.2/39.6 & 45.2/29.6 & 60.7  \\ 
$\joint{}_{\text{250K}}$ & 265M &1.00x & 69.2 & 83.3 & 72.4 & 59.7 & 63.9/47.9 & 58.9/40.7 & 45.4/29.6 & 61.1 \\ 
$\joint{}_{\text{500K}}$ & 448M &0.72x & 69.4 & 82.2 & 72.1 & 60.5 & 64.7/48.0 & 58.2/40.3 & 48.0/32.6 & 61.4 \\ 
$\ours{}_{\text{250K}}$ & 265M &1.00x & 69.3 & 82.0 & 71.4 & 60.0 & 66.2/50.3 & 60.1/42.6 & 45.6/30.6 & 61.5 \\ 
$\ours{}_{\text{500K}}$ & 448M &0.72x & 70.5 & 83.0 & \textbf{72.9} & \textbf{62.7} & 66.8/\textbf{50.6} & 60.9/\textbf{42.9} & 50.0/34.5 & 63.1 \\
~~+ $k$-NN & 448M &1.18x & \textbf{70.8} & 82.6 & 72.5 & 61.8 & \textbf{67.1}/49.8 & \textbf{61.4}/42.5 & \textbf{56.3/39.3} & \textbf{63.7} \\ 
~~+ half emb & 265M & 0.94x & 70.3 & 83.0 & 72.0 & 61.7 & 65.8/49.0 & 61.0/42.3 & 49.3/33.0 & 62.5 \\ 
~~+ $k$-NN \& half emb  & 265M & \textbf{1.35x} & 69.8 & \textbf{83.4} & 72.1 & 60.1 & 66.6/49.5 & 60.8/42.7 & 50.2/33.9 & 62.5 \\ 
\bottomrule
\end{tabular}
\caption{Evaluation results on the XTREME benchmark. ``$\text{XLM-R}_{\text{250K}}$'' denotes using the XLM-R~\citep{xlmr} vocabulary with 250K subword units. ``$k$-NN'' and ``half emb'' denote our $k$-NN-based target sampling method and using half embedding dimension, respectively. }
\label{table:benchmark-results}
\end{table*}

\subsection{Reducing the Embedding Dimension}
\label{sec:mb}

In order to keep the number of model parameters fixed while increasing the vocabulary size, we follow \cite{albert} and \cite{rembert} to reduce both the input and output embedding dimension and linearly project the embeddings to the hidden dimension of the Transformer blocks. 
More precisely, we halve the embedding dimension when the vocabulary size is doubled. 
This rebalancing strategy only slightly degrades the model performance but improves pre-training speed and decreases the model size.

\citet{xlmr} also studied the relation between the size of the shared multilingual vocabulary and downstream task performance with multilingual models of the fixed number of parameters.
They keep the overall number of parameters constant by adjusting the width (i.e., hidden size) of the Transformer. 
Notice that we only reduce the embedding dimension while keeping the Transformer blocks untouched.

\section{Experiments}

\subsection{Setup}

\paragraph{Fine-Tuning Datasets} 
To validate the effectiveness of our methods, we conduct experiments on three types of cross-lingual understanding tasks from XTREME benchmark~\cite{xtreme}, including two classification datasets: XNLI~\cite{xnli}, PAWS-X~\cite{pawsx}, three span extraction datasets: XQuAD~\cite{xquad}, MLQA~\cite{mlqa}, TyDiQA-GoldP~\cite{tydiqa}, and two sequence labeling datasets: NER~\cite{panx}, POS~\cite{udpos}. 
The statistics of the datasets are shown in the appendix.

\paragraph{Implementation Details}
We adapt the Transformer architecture from the base model setting in \citet{xlmr}, i.e., 12 layers and 768 hidden dimension size.
We use masked language modeling objective to train our models for 1 million updates on eight 32GB Nvidia V100 GPUs with a batch size of 256. 
We update the top-k indices for every word in the multilingual vocabulary every 1,000 training steps and use $k=50$ in $k$-NN-based target sampling.
The learning rate is scheduled with a polynomial decay with 10K warmup steps, where the peak learning rate is set as 0.0001. We adapt other hyper-parameters in pre-training from~\citet{infoxlm}.
All fine-tuning results are averaged over five random seeds.
The fine-tuning pipeline is based on the code base of~\cite{xtune}.
The fine-tuning implementation details are shown in the appendix.

\subsection{Results}
Table~\ref{table:benchmark-results} shows XTREME fine-tuning results with models pretrained using different vocabularies and acceleration strategies. 
Compared to vocabulary directly learned on multilingual corpus with SentencePiece, i.e., $\text{XLM-R}_{\text{250K}}$ and $\joint{}_{\text{250K}}$, our $\ours{}_{\text{250K}}$ improves on question answering datasets but degrades on PAWS-X, POS and NER.
Then increasing the vocabulary from $\ours{}_{\text{250K}}$ to $\ours{}_{\text{500K}}$ mitigates the gap and bring improvements on six datasets except for PAWS-X, which only includes seven high-resource languages.
However, increasing the size of vocabulary directly learned with Sentencepiece from $\joint{}_{\text{250K}}$ to $\joint{}_{\text{500K}}$ does not improve the performance as our \ours{} method does, showing the importance of selecting language-specific subword units and leveraging how much vocabulary capacity each language requires.

Since increasing vocabulary size brings the issues of model size and pre-training speed, we study the proposed method to accelerate pre-training: $k$-NN-based target sampling ($k$-NN) and using half embedding dimension (half emb).
Our $k$-NN method improves pre-training speed with a 500K vocabulary so that the speed is 1.18 times that vanilla pre-training with a 250K vocabulary. Meanwhile, pre-training with our $k$-NN method does not significantly degrade the performance, it even brings improvement on XNLI, MLQA, and TyDiQA. 
Then we halve the embedding dimension of the models with 500K vocabulary and results in a similar number of parameters to models with 250K vocabulary. 
The overall performance degrades by 0.6-points but still consistently improves over models with 250K vocabularies while the speed is comparable.
Combining the two methods above, we achieve a 1.35-times speed-up and more than 1 point improvement with a similar model size compared to models with 250K vocabularies.

\subsection{Analysis and Discussion}
We conduct a thorough analysis to understand the impact of our proposed methods on cross-lingual language models.  
To reduce the computation load, we only pre-train the cross-lingual language models for 500K steps for some of our settings.

\begin{table}[t]
\centering
\small
\begin{tabular}{lcccc}
\toprule
\textbf{Method} & \textbf{XNLI} & \textbf{POS} & \textbf{MLQA} & \textbf{Speed} \\ \midrule
$\ours{}_{\text{500K}}$ & 69.2 & \textbf{72.9} & \textbf{59.9/41.7} & 1.00x \\ \midrule
+ $k$-NN & \textbf{69.3} & 72.1 & 59.6/40.3 & \textbf{1.64x} \\
+ target sampling & 68.8 & 71.3 & 57.6/38.8 & 1.56x \\
+ NCE & 56.0 & 61.8 & 41.1/26.2 & 1.40x \\
+ NEG & 56.5 & 62.9 & 40.1/25.6 & 1.40x \\ 
\bottomrule
\end{tabular}
\caption{Comparison between different sampling-based softmax approximation approaches with vocabulary $\ours{}_{\text{500K}}$.  Models are pretrained for 0.5M steps.}
\label{table:sampling-results}
\end{table}

\begin{table}[t]
\centering
\small
\begin{tabular}{lcccc}
\toprule
\textbf{Method} & \textbf{XNLI} & \textbf{POS}  & \textbf{MLQA} & \textbf{Speed}\\ \midrule
$\ours{}_{\text{500K}}$ & 69.2 & 71.8 & 59.9/\textbf{41.7} & 1.00x \\ \midrule
+ $k$-NN ($k$=5) & 68.5 & 71.3 & 58.6/40.0 & \textbf{1.76x} \\
+ $k$-NN ($k$=10) & \textbf{69.3} & 71.4 & 58.9/39.6 & 1.74x \\
+ $k$-NN ($k$=25) & 69.2 & 71.7 & 59.8/40.9 & 1.69x \\
+ $k$-NN ($k$=50) & \textbf{69.3} & \textbf{72.1} & 59.6/40.3 & 1.64x \\
+ $k$-NN ($k$=100) & 69.5 & \textbf{72.1} & \textbf{60.0}/41.3 & 1.57x \\
\bottomrule
\end{tabular}
\caption{Comparison between different $k$ values in $k$-NN-based sampling method. Models are pretrained for 0.5M steps.}
\label{table:k-value}
\end{table}

\paragraph{$k$-NN-based target sampling outperforms previous sampling-based approaches.} 
To verify the effectiveness of our proposed $k$-NN-based sampling method, we compare it against previous sampling-based approaches used to approximate softmax, which are target sampling~\cite{DBLP:conf/acl/JeanCMB15}, noise contrastive estimation (\citet{nce}, NCE) and negative sampling (\citet{word2vec}, NEG). The results are shown in Table~\ref{table:sampling-results}. 
To make a fair comparison, since our $k$-NN-based sampling method using $k=50$ samples vocabulary subset with less than 50,000 subword units per batch on average, we here sample 50,000 negative subword units per batch for target sampling, NCE, and NEG. 
Among the four methods, NCE and NEG are significantly worse than $k$-NN and target sampling. We attribute it that NCE and NEG need more training steps to converge~\citep{nce}.  
Besides, the original NCE typically sample different negative samples for every target word, while we here use 50,000 negative samples for all target word in current mini-batch, which is more efficient on GPUs.

\paragraph{Effect of the value of $k$ in $k$-NN-based target sampling.}
We illustrate the downstream task performance when using different values of $k$ in our $k$-NN-based target sampling in Table~\ref{table:k-value}. 
While a smaller $k$ indicates faster pre-training speed, we observe even with a small value like 5, the result does not significantly degrade compared to using the original softmax. 
We attribute this to that by retrieving subword samples that are most similar to the target subword, the model can focus on the difficult part of the original masked language modeling objective.
More precisely, the model focus on discriminating the ground-truth subword from a set of noise samples that are not easy to distinguish. 
Considering the overall performance, the pre-training speed, and running memory to store $k$-NN indices, we use $k=50$ in all our experiments.

\begin{table}[t]
\centering
\small
\begin{tabular}{lcccc}
\toprule
\textbf{Method} & \textbf{XNLI} & \textbf{POS} & \textbf{NER} & \textbf{MLQA}\\ \midrule
$\beta$=0 & 66.9 & \textbf{71.8} & 61.5 & 58.6/41.0 \\ 
$\beta$=0.3 & 69.0 & 71.7 & \textbf{61.6} & 59.2/40.1 \\ 
$\beta$=0.7 & 69.2 & \textbf{71.8} & 61.5 & \textbf{59.9/41.7} \\ 
$\beta$=1.0 & \textbf{69.5} & \textbf{71.8} & 60.9 & 58.4/40.3 \\ 
\bottomrule
\end{tabular}
\caption{Impact of adjusting high-resource versus low-resource vocabulary capacity trade-off with $\beta$. 
$\beta=0$ indicates the vocabulary is allocated without considering pre-training corpus size. Models are pretrained for 0.5M steps.}
\label{table:beta}
\end{table}

\paragraph{Language-specific pre-training corpus should also be considered when allocating vocabulary capacity.}
The pre-training corpus size varies across different languages.
It is inefficient to allocate a large vocabulary capacity for low-resource languages with rare pre-training data since the pretrained model can only learn limited knowledge from these languages. 
Here we study the value of rescaling factor $\beta$ from \eqform{eq:vocap} in multilingual vocabulary construction in Table~\ref{table:beta}. 
The rescaling factor $\beta$ controls the number of selected language-specific subword units. 
Increasing the value of $\beta$ improves the performance of XNLI, where most languages are high-resource languages. 
However, it degrades the performance of NER, where more low-resources languages exist.
When considering overall performance, we decide to use $\beta=0.7$ in our experiments.

\begin{figure}[t]
\centering
\includegraphics[trim={2.4cm 0cm 4cm 1.4cm},scale=0.175,clip]{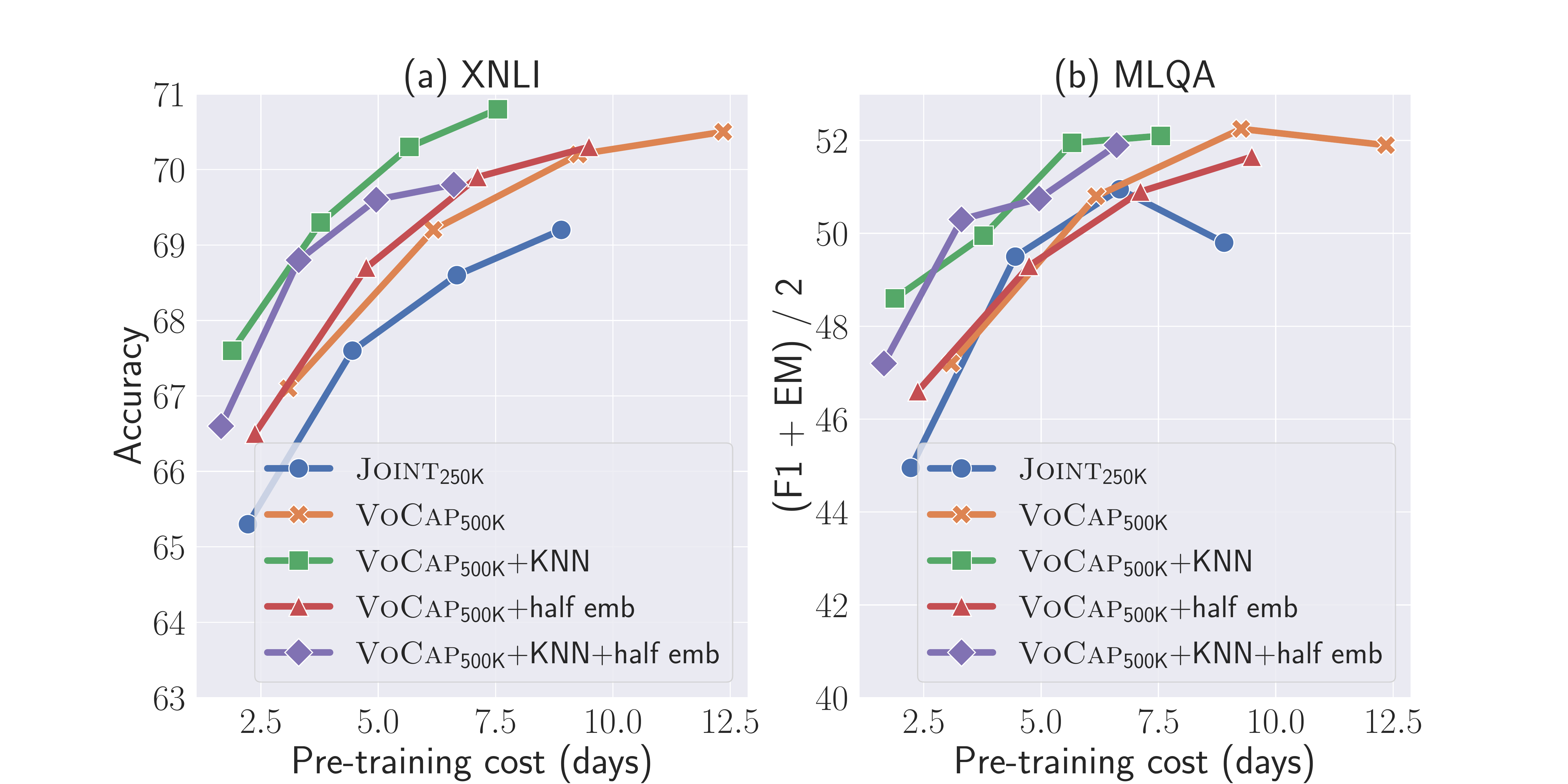}
\caption{Performance on XNLI and MLQA versus the cross-lingual language models' pre-training cost.}
\label{fig:time-performance}
\end{figure}

\paragraph{The proposed acceleration strategies significantly improve the downstream task performance under the same pre-training cost.} 
Increasing the vocabulary size slows the pre-training speed, even though there is almost no difference in fine-tuning speed. 
We study the relationship between the downstream task performance and the pre-training cost under different model settings in Figure~\ref{fig:time-performance}. 
We observe $\ours{}_{\text{500K}}$+$k$-NN achieves the best performance.
Models trained with 500K vocabulary consistently outperform 250K vocabulary on XNLI. 
Besides, we observe the performance on MLQA with the model trained using 250K vocabulary degrades as the training continues while models trained using 500K vocabulary does not, indicating the sufficient vocabulary capacity is essential for question answering task.

\begin{figure}[t]
\centering
\includegraphics[trim={1.2cm 0cm 2cm 0.7cm},scale=0.35,clip]{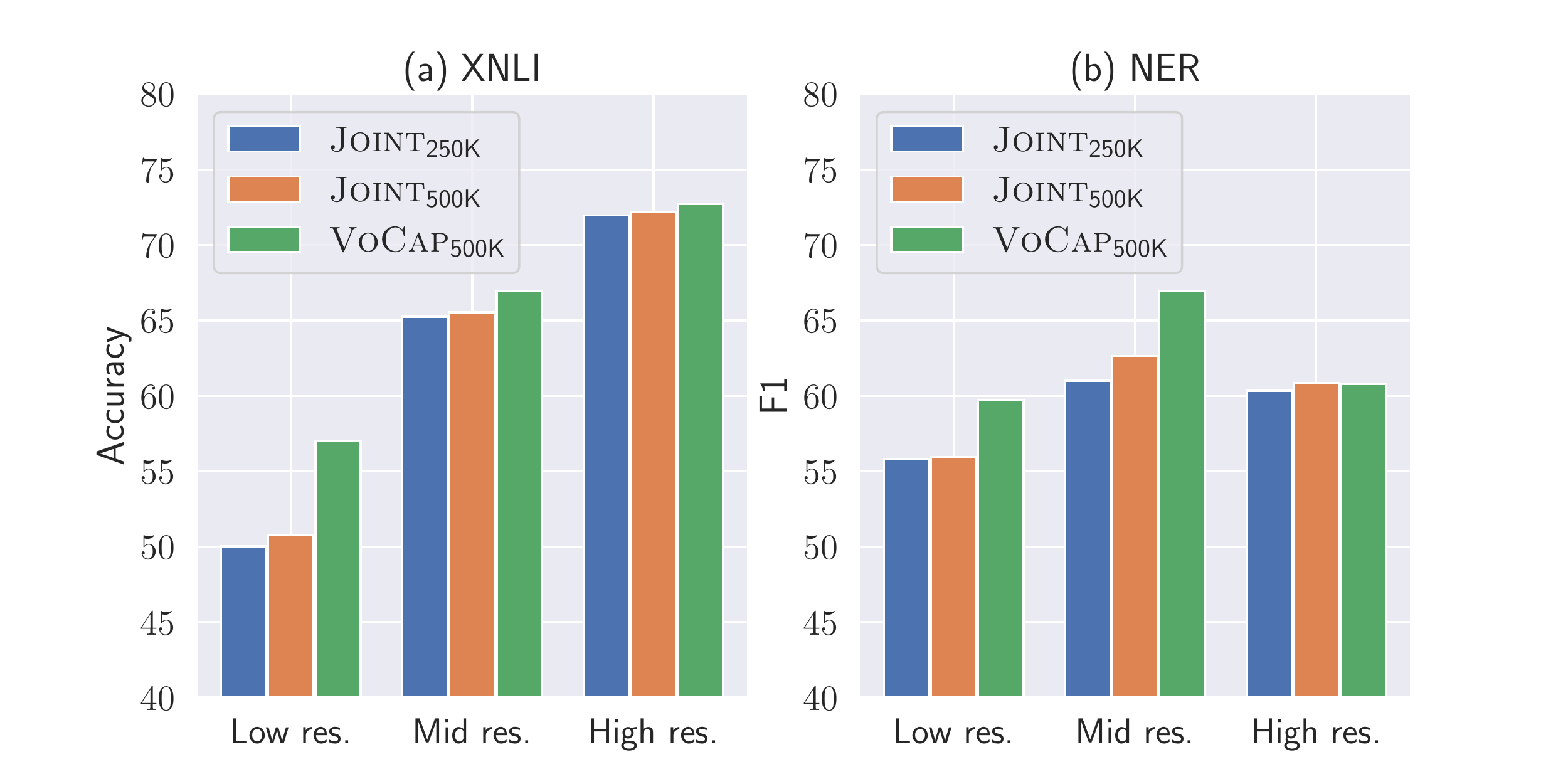}
\caption{Impact of \ours{} on the performance of different-resourced languages on XNLI and NER.}
\label{fig:res-performance}
\end{figure}

\paragraph{\ours{} gains more improvement on mid and low-resource languages than high-resource languages.
} 
In Figure~\ref{fig:res-alp} in Section~\ref{sec:vocap}, we show that the vocabulary learned with \ours{} benefits the vocabulary capacity of low-resource languages more than high-resource languages, indicating the improvements should mainly come from low-resource languages. 
To verify this, we compare \ours{} against SentencePiece baseline on the performance of different-resourced languages on XNLI and NER in Figure~\ref{fig:res-performance}.
We observe that the vocabulary learned with \ours{} significantly outperforms the vocabularies directly learned with SentencePiece on mid and low-resource languages. This observation is also consistent with the ALP results in Figure~\ref{fig:res-alp}.

\section{Related Work}

\paragraph{Pretrained Cross-Lingual Language Models} 
Recent work pre-trains Transformer models~\citep{transformer} on the large-scale multilingual corpus to obtain pretrained cross-lingual language models~\citep{xlm,xlmr,xnlg,mt6,infoxlm,xlmalign,xlme,rembert,mt5,xlmt,deltalm}.
These models are capable of encoding texts from different languages into universal representations and significantly improves cross-lingual transferability.

\paragraph{Multilingual Vocabulary Construction} 
Cross-lingual language models need large vocabularies to ensure all languages are adequately represented.
Recent research work on constructing multilingual vocabulary for cross-lingual language models can be categorized into two groups. mBERT~\citep{bert}, XLM~\citep{xlm}, and XLM-R~\citep{xlmr} learn vocabularies on a combined multilingual corpus with WordPiece~\citep{wordpiece}, BPE~\citep{bpe}, and unigram language model~\citep{DBLP:conf/acl/Kudo18} from SentencePiece~\citep{sentencepiece}, respectively.
\citet{DBLP:conf/emnlp/ChungGTR20} propose to balance the trade-off between optimizing for cross-lingual subword sharing and the need for robust representation of individual languages. They first group languages into clusters and learn vocabularies individually on each cluster, then combine all cluster-vocabularies to form a single unified multilingual vocabulary. 
Compared to \citet{DBLP:conf/emnlp/ChungGTR20}, our advantage is that we separately quantify the vocabulary capacity each language needs with average log probability and balance the construction procedure with pre-training corpus size.

\paragraph{Softmax Approximation}
Approximating the softmax was a core problem in training NLP tasks with a large vocabulary, e.g., neural machine translation, language modeling. 
With the rise of subword representations~\citep{bpe,wordpiece,DBLP:conf/acl/Kudo18}, the vocabulary size significantly decreases, and the problem has been less studied recently.
Nevertheless, the need for training cross-lingual language models with a large multilingual vocabulary has drawn our attention again to the softmax approximation approaches.
The existing softmax approximation approaches can be grouped into softmax-based and sampling-based approaches. Softmax-based approaches includes hierarchical softmax~\citep{h-softmax}, differentiated softmax~\citep{d-softmax}, and CNN-softmax~\citep{cnn-softmax}. 
However, these approaches improve the softmax efficiency by changing its architecture, which is unsuitable for either training on GPUs or multilingual settings.
Sampling-based approaches instead optimize some other easy-to-compute loss function to approximate the original softmax, including target sampling~\citep{DBLP:conf/acl/JeanCMB15}, noise contrastive estimation~\citep{nce}, negative sampling~\citep{word2vec}.
Our $k$-NN-based target sampling is also a sampling-based approach.

\section{Conclusion}

In this paper, we study pre-training cross-lingual language models with large vocabulary capacity. First, we propose \ours{} to construct large multilingual vocabulary in cross-lingual language models. 
We conduct a quantitative analysis to show that average log probability is an valid indicator of vocabulary capacity for a particular language, which also correlates with downstream task performance on the language.
\ours{} uses the language-specific average log probability and pre-training corpus size to allocate appropriate vocabulary capacity for each language in the multilingual vocabulary.
Moreover, we propose $k$-NN-based target sampling to accelerate pre-training with the allocated large multilingual vocabulary by approximating the expensive softmax.
We also show that reducing the embedding dimension is an effective way to keep the improvement brought by the large vocabulary without increasing the number of model parameters.
The experiments demonstrate the effectiveness of the proposed vocabulary construction method as well as the acceleration methods.

\bibliographystyle{acl_natbib}
\bibliography{vocap}

\appendix

\section{Correlation between Language-Specific Vocabulary Capacity and Task Performance}
We compare the Pearson correlation coefficients between ALP and downstream task performance with the coefficients between vocabulary size and downstream task performance in Table~\ref{table:coeff}. The results show that ALP correlates better than vocabulary size with downstream task performance.

\begin{table}[ht]
\begin{tabular}{lccc}
\toprule
Language & Task & $\rho(\text{ALP}, \text{F1})$ & $\rho(|V|, \text{F1})$ \\
\midrule
\multirow{2}{*}{hi} & POS & \textbf{0.922} & 0.787   \\
 & NER & 0.879 & \textbf{0.890}   \\
\midrule
\multirow{2}{*}{fa} & POS & \textbf{0.905}            &  0.700  \\
 & NER & \textbf{0.912}            &  0.872  \\
\midrule
\multirow{2}{*}{it} & POS & \textbf{0.665}            &  0.422  \\
 & NER & 0.899            &  \textbf{0.900}  \\
\midrule
\multirow{2}{*}{ru} & POS & \textbf{0.423}            &  0.327  \\
 & NER & \textbf{0.872}            &  0.833  \\
\bottomrule
\end{tabular}
\caption{Pearson correlation coefficients between ALP and downstream task performance and between vocabulary size and downstream task performance.}
\label{table:coeff}
\end{table}

\section{Statistics of XTREME Datasets}
\begin{table}[ht]
\centering
% \small
\begin{tabular}{llll}
\toprule
Task                                                  & Dataset & $| \text{Train} |$ & $| \text{Lang} |$ \\ \midrule
\multirow{2}{*}{Classification}                       & XNLI    & 392K                  & 15                      \\
& PAWS-X  & 49.4K                 & 7                       \\ \midrule
Structured                                            & POS     & 21K                   & 33                      \\
Prediction                                            & NER     & 20K                   & 40                      \\ \midrule
\multirow{3}{*}{\tabincell{c}{Question \\ Answering}} & XQuAD   & 87K                   & 11                      \\
& MLQA    & 87K                   & 7                       \\
& TyDiQA  & 3.7K                  & 9                       \\ \bottomrule
\end{tabular}
\caption{Statistics for the datasets in the XTREME benchmark. we report the number of training examples ($| \text{Train} |$), and the number of languages ($| \text{Lang} |$).}
\label{table:dataset}
\end{table}

\begin{table}[ht]
\scriptsize
\centering
\begin{tabular}{crcrcr}
\toprule 
Code & Size (GB) & Code & Size (GB) & Code & Size (GB) \\
\cmidrule(r){1-2}  \cmidrule(r){3-4} \cmidrule(r){5-6}
af & 0.2 & hu & 9.5 & pl & 28.6 \\
am & 0.4 & hy & 0.7 & ps & 0.4 \\
ar & 16.1 & id & 17.2 & pt & 39.4 \\
as & 0.1 & is & 0.5 & ro & 11.0 \\
az & 0.8 & it & 47.2 & ru & 253.3 \\
ba & 0.2 & ja & 86.8 & sa & 0.2 \\
be & 0.5 & ka & 1.0 & sd & 0.2 \\
bg & 7.0 & kk & 0.6 & si & 1.3 \\
bn & 5.5 & km & 0.2 & sk & 13.6 \\
ca & 3.0 & kn & 0.3 & sl & 6.2 \\
cs & 14.9 & ko & 40.0 & sq & 3.0 \\
cy & 0.4 & ky & 0.5 & sr & 7.2 \\
da & 6.9 & la & 0.3 & sv & 60.4 \\
de & 99.0 & lo & 0.2 & sw & 0.3 \\
el & 13.1 & lt & 2.3 & ta & 7.9 \\
en & 731.6 & lv & 1.3 & te & 2.3 \\
eo & 0.5 & mk & 0.6 & tg & 0.7 \\
es & 85.6 & ml & 1.3 & th & 33.0 \\
et & 1.4 & mn & 0.4 & tl & 1.2 \\
eu & 1.0 & mr & 0.5 & tr & 56.4 \\
fa & 19.0 & ms & 0.7 & tt & 0.6 \\
fi & 5.9 & mt & 0.2 & ug & 0.2 \\
fr & 89.9 & my & 0.4 & uk & 13.4 \\
ga & 0.2 & ne & 0.6 & ur & 3.0 \\
gl & 1.5 & nl & 25.9 & uz & 0.1 \\
gu & 0.3 & nn & 0.4 & vi & 74.5 \\
he & 4.4 & no & 5.5 & yi & 0.3 \\
hi & 5.0 & or & 0.3 & zh & 96.8 \\
hr & 1.4 & pa & 0.8 \\
\bottomrule 
\end{tabular}
\caption{The statistics of the reconstructed CommonCrawl corpus for learning vocabularies.}
\label{table:ccnet}
\end{table}

\begin{table}[ht]
\scriptsize
\centering
\begin{tabular}{crcrcr}
\toprule 
Code & Size (GB) & Code & Size (GB) & Code & Size (GB) \\
\cmidrule(r){1-2}  \cmidrule(r){3-4} \cmidrule(r){5-6}
af & 0.12 & hu & 0.8 & pl & 1.55 \\
am & 0.01 & hy & 0.6 & ps & 0.04 \\
ar & 1.29 & id & 0.52 & pt & 1.5 \\
as & 0.04 & is & 0.05 & ro & 0.42 \\
az & 0.24 & it & 2.69 & ru & 5.63 \\
ba & 0.13 & ja & 2.65 & sa & 0.04 \\
be & 0.31 & ka & 0.37 & sd & 0.02 \\
bg & 0.62 & kk & 0.29 & si & 0.09 \\
bn & 0.41 & km & 0.12 & sk & 0.21 \\
ca & 1.1 & kn & 0.25 & sl & 0.21 \\
cs & 0.8 & ko & 0.56 & sq & 0.1 \\
cy & 0.06 & ky & 0.1 & sr & 0.74 \\
da & 0.33 & la & 0.05 & sv & 1.7 \\
de & 5.43 & lo & 0.01 & sw & 0.03 \\
el & 0.73 & lt & 0.19 & ta & 0.46 \\
en & 12.58 & lv & 0.12 & te & 0.44 \\
eo & 0.25 & mk & 0.34 & tg & 0.04 \\
es & 3.38 & ml & 0.28 & th & 0.52 \\
et & 0.23 & mn & 0.05 & tl & 0.04 \\
eu & 0.24 & mr & 0.1 & tr & 0.43 \\
fa & 0.66 & ms & 0.2 & tt & 0.09 \\
fi & 0.68 & mt & 0.01 & ug & 0.03 \\
fr & 4.0 & my & 0.15 & uk & 2.43 \\
ga & 0.03 & ne & 0.06 & ur & 0.13 \\
gl & 0.27 & nl & 1.38 & uz & 0.06 \\
gu & 0.09 & nn & 0.13 & vi & 0.76 \\
he & 1.11 & no & 0.54 & yi & 0.02 \\
hi & 0.38 & or & 0.04 & zh & 1.08 \\
hr & 0.28 & pa & 0.1 & \\
\bottomrule 
\end{tabular}
\caption{The statistics of the Wikipedia corpus used for pre-training.}
\label{table:wikipedia}
\end{table}.

\begin{table}[ht]
\scriptsize
\centering
\begin{tabular}{crcrcr}
\toprule 
Code & Size (K) & Code & Size (K) & Code & Size (K) \\
\cmidrule(r){1-2}  \cmidrule(r){3-4} \cmidrule(r){5-6}
af & 2 & hu & 12 & pl & 20 \\
am & 3 & hy & 5 & ps & 3 \\
ar & 15 & id & 13 & pt & 20 \\
as & 2 & is & 3 & ro & 13 \\
az & 5 & it & 22 & ru & 34 \\
ba & 2 & ja & 23 & sa & 1 \\
be & 3 & ka & 4 & sd & 2 \\
bg & 9 & kk & 4 & si & 3 \\
bn & 6 & km & 4 & sk & 11 \\
ca & 8 & kn & 2 & sl & 8 \\
cs & 14 & ko & 17 & sq & 7 \\
cy & 3 & ky & 3 & sr & 10 \\
da & 9 & la & 3 & sv & 18 \\
de & 24 & lo & 2 & sw & 3 \\
el & 17 & lt & 7 & ta & 6 \\
en & 23 & lv & 6 & te & 4 \\
eo & 4 & mk & 4 & tg & 5 \\
es & 26 & ml & 3 & th & 14 \\
et & 5 & mn & 3 & tl & 4 \\
eu & 4 & mr & 3 & tr & 18 \\
fa & 9 & ms & 4 & tt & 3 \\
fi & 9 & mt & 3 & ug & 3 \\
fr & 25 & my & 2 & uk & 12 \\
ga & 2 & ne & 3 & ur & 5 \\
gl & 5 & nl & 14 & uz & 2 \\
gu & 2 & nn & 3 & vi & 12 \\
he & 6 & no & 7 & yi & 2 \\
hi & 6 & or & 2 & zh & 30 \\
hr & 6 & pa & 3 & \\
\bottomrule 
\end{tabular}
\caption{The statistics of the allocated vocabulary size for each language.}
\label{table:vocab-size}
\end{table}

\section{Fine-tuning Settings}
\paragraph{Implementation Details}
For the POS dataset, we use the average-pooling strategy on subwords to obtain word representation since part-of-speech is related to different parts of words, depending on the language. 
We tune the hyper-parameter and select the model with the best average results over all the languages' development set. 
There are two datasets without development set in multi-languages.
For XQuAD, we tune the hyper-parameters with the development set of MLQA since they share the same training set and have a higher degree of overlap in languages. For TyDiQA-GoldP, we use the English test set as the development set. 
\paragraph{Hyper-Parameters}
For XNLI, PAWS-X, POS, and NER, we fine-tune 10 epochs.
For XQuAD and MLQA, we fine-tune 4 epochs. For TyDiQA-GoldP, we fine-tune 6 or 8 epochs and select the best number of epochs with the English test set as the development set.  
For learning rate, we select in [7e-6, 1e-5] for XNLI and PAWS-X, [1e-5, 2e-5] for POS and NER, [2e-5, 3e-5] for XQuAD, MLQA and TyDiQA-GoldP.
\section{Pre-Training Data}
We use the reconstruct CommonCrawl corpus in \citet{infoxlm} to learn vocabularies in our paper.
Because tokenizing the pre-training data is time-consuming, we instead conduct our pre-training on Wikipedia since it has a smaller size. 
We only consider the languages that are shared by the reconstructed CommonCrawl corpus and Wikipedia.
The statistics of the Wikipedia corpus and the reconstructed CommonCrawl corpus are listed in Table~\ref{table:wikipedia} and Table~\ref{table:ccnet}.

\end{document}

%% file: settings.tex
\usepackage{multirow}
\usepackage{amsmath}
\usepackage{capt-of}
\usepackage{tabularx}
\usepackage{epsfig}
\usepackage{amssymb}
\usepackage{amsfonts}
\usepackage{booktabs}
\usepackage{scalerel}
\usepackage[inline]{enumitem}
\usepackage{listings}
\usepackage{varwidth}
\usepackage[export]{adjustbox}
\usepackage{tikz}
\usetikzlibrary{tikzmark}
\usepackage{cleveref}

\usepackage{stmaryrd}
\usepackage{bbm}

\newcommand{\tabincell}[2]{\begin{tabular}{@{}#1@{}}#2\end{tabular}}

\newcommand{\eqform}[1]{Equation~(\ref{#1})}

\usepackage{algorithm}
\usepackage[noend]{algpseudocode}

\definecolor{deepblue}{rgb}{0,0,0.5}
\definecolor{officeblue}{RGB}{0,102,204}
\definecolor{deepred}{rgb}{0.6,0,0}
\definecolor{deepgreen}{rgb}{0,0.5,0}
\definecolor{mybrickred}{RGB}{182,50,28}

\definecolor{fillcolor}{RGB}{216,217,252}

\algnewcommand\algorithmicrequireb{{\hspace{0.85cm}}}
\algnewcommand\INPTDESCB{\item[\algorithmicrequireb]}

\algnewcommand\algorithmicfuncdesc{\textbf{Function:}}
\algnewcommand\FUNCDESC{\item[\algorithmicfuncdesc]}
\algnewcommand\algorithmicfuncdescb{{\hspace{1.48cm}}}
\algnewcommand\FUNCDESCB{\item[\algorithmicfuncdescb]}
\algnewcommand{\algorithmicgoto}{\textbf{goto}}
\algnewcommand{\Goto}[1]{\algorithmicgoto~\ref{#1}}

%% file: math_commands.tex
\usepackage{amsmath,amsfonts,bm}

\def\eqref#1{equation~\ref{#1}}
\def\1{\bm{1}}
\newcommand{\train}{\mathcal{D}}

\def\vtheta{{\bm{\theta}}}

\def\vg{{\bm{g}}}

\DeclareMathAlphabet{\mathsfit}{\encodingdefault}{\sfdefault}{m}{sl}
\SetMathAlphabet{\mathsfit}{bold}{\encodingdefault}{\sfdefault}{bx}{n}

\newcommand{\Ls}{\mathcal{L}}